\documentclass[10pt,twocolumn,letterpaper]{article}

\usepackage{iccv}
\usepackage{times}
\usepackage{epsfig}
\usepackage{graphicx}
\usepackage{amsmath}
\usepackage{amssymb}
\usepackage{diagbox}
\usepackage{xcolor}
\usepackage{subcaption}
\usepackage{cite}
\usepackage{float,lscape}
\usepackage{authblk}
\usepackage{algorithmicx}
\usepackage{algpseudocode}
\usepackage{algorithmicx}
\usepackage{algorithm}
\usepackage{rotating}
\usepackage{adjustbox}
\usepackage{blindtext}
\iccvfinalcopy 

\def\iccvPaperID{10957} 

\ificcvfinal\fi 

\definecolor{citecolor}{HTML}{0071bc}
\usepackage[pagebackref,breaklinks=true,letterpaper=true,colorlinks,citecolor=citecolor,bookmarks=false]{hyperref}

\title{Anomaly Detection under Distribution Shift}
\author{Tri Cao}
\author{Jiawen Zhu}
\author{Guansong Pang\thanks{Corresponding author: G. Pang ({\tt pangguansong@gmail.com}).}}
\affil{School of Computing and Information Systems, Singapore Management University}
\begin{document}
\maketitle
\begin{abstract}

Anomaly detection (AD) is a crucial machine learning task that aims to learn patterns from a set of normal training samples to identify abnormal samples in test data. 
Most existing AD studies
assume that the training and test data are drawn from the same data distribution, but the test data can have large distribution shifts arising in many real-world applications due to different natural variations such as new lighting conditions, object poses, or background appearances, rendering existing AD methods ineffective in such cases.
In this paper, we consider the problem of anomaly detection under distribution shift and establish performance benchmarks on four widely-used AD and out-of-distribution (OOD) generalization datasets. We demonstrate that simple adaptation of state-of-the-art OOD generalization methods to AD settings fails to work effectively due to the lack of labeled anomaly data. We further introduce a novel robust AD approach to diverse distribution shifts by minimizing the distribution gap between in-distribution and OOD normal samples in both the training and inference stages in an unsupervised way. Our extensive empirical results on the four datasets show that our approach substantially outperforms state-of-the-art AD methods and OOD generalization methods on data with various distribution shifts, while maintaining the detection accuracy on in-distribution data. Code and data are available at \renewcommand\UrlFont{\color{blue}\tt}
\url{https://github.com/mala-lab/ADShift}.

\end{abstract}
\section{Introduction}


\begin{figure}[t]
    \centering
    \includegraphics[width=0.5\textwidth]{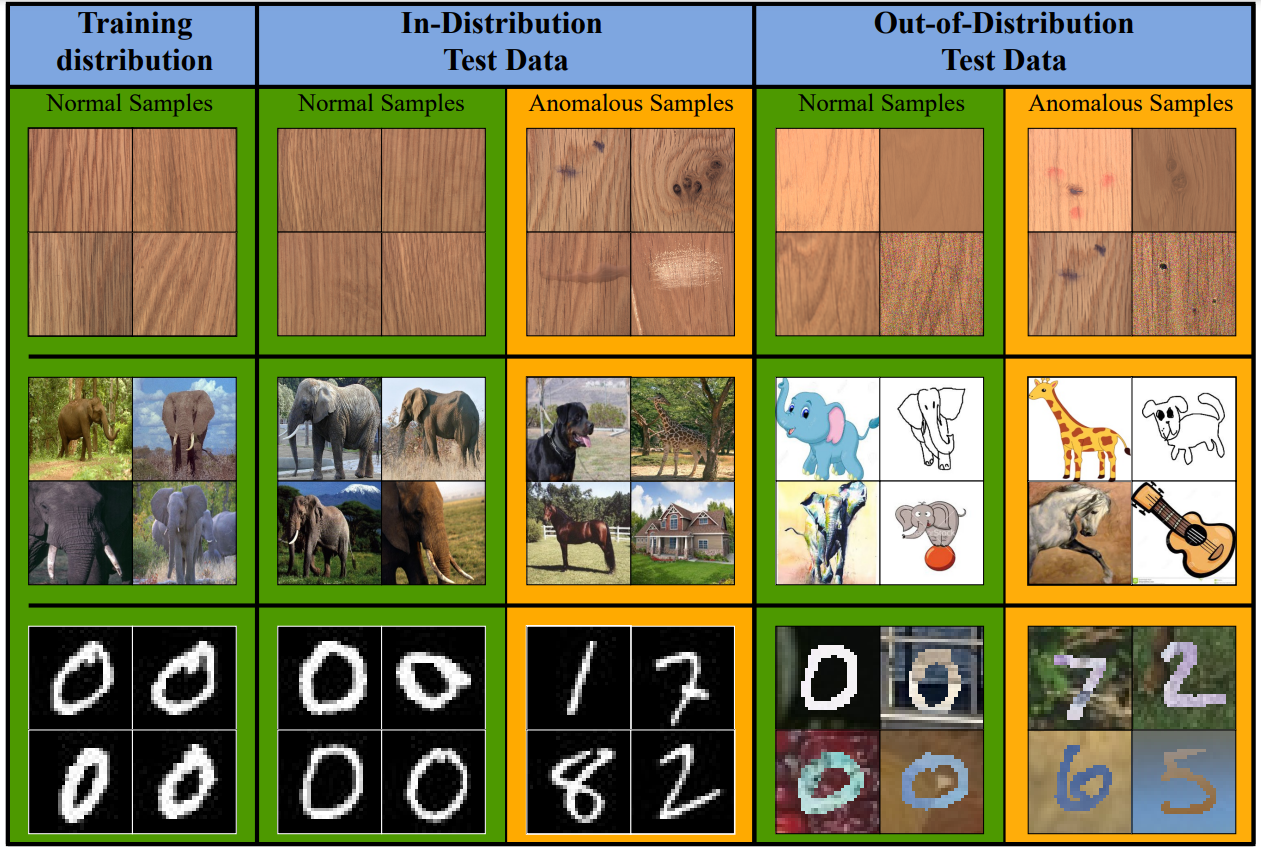}
    \caption{Illustrative samples for anomaly detection under distribution shift. First row: the `Wood' dataset from MVTec \cite{bergmann2019mvtec}. Second row: the `Elephant' class as normal and the remaining classes as anomaly in PACS \cite{li2017deeper}. Third row: the `0' class as normal and the remaining classes as anomaly in MNIST\cite{lecun1998gradient}/MNIST-M\cite{ganin2015unsupervised}. We aim at distinguishing anomalies from normal data in both in-distribution test data and out-distribution test data
    }
    \label{fig:intro}
\end{figure}
Anomaly Detection (AD) is a crucial task in machine learning that aims to identify rare and unusual patterns in data. It is an important problem in various domains, such as financial domain \cite{ahmed2016survey, anandakrishnan2018anomaly}, cybersecurity \cite{ten2011anomaly, siddiqui2019detecting}, industrial inspection \cite{bergmann2019mvtec}, and medical diagnosis \cite{schlegl2019f, shvetsova2021anomaly}. 
Due to the difficulty and/or high cost of collecting labeled anomaly data,
current AD studies are focused on unsupervised approaches, which aim to learn patterns from a set of normal training samples to identify abnormal samples in test data.

Although existing AD studies have demonstrated promising performance \cite{li2021cutpaste, salehi2021multiresolution, deng2022anomaly,pang2021deep}, they generally assume that the training and test data are drawn from the same data distribution. However, this assumption is often unrealistic in real-world scenarios as the test data can have large distribution shifts arising in many applications due to different natural variations such as new lighting conditions, object poses, or background appearances, rendering the AD methods ineffective in such cases.

Distribution shift is a ubiquitous problem in different real-world applications, which can significantly degrade the performance of models in various tasks such as image classification, object detection, and segmentation \cite{koh2021wilds, ganin2015unsupervised, li2017deeper, zhang2020generalizing}. Many out-of-distribution (OOD) generalization methods have been introduced to address this problem \cite{huang2020self, hendrycks2019augmix, zhang2022exact, carlucci2019domain, ghifary2015domain, lehner20223d, ma2020training, liu2021feddg, ouyang2022causality}. 
These OOD generalization methods rely on large labeled training data from one or multiple relevant domains to learn domain-invariant feature representations. They often require class labels \cite{yao2022pcl, huang2020self, qiao2020learning}, domain labels \cite{zhao2021learning, yoon2019generalizable, wang2020cross, carlucci2019hallucinating}, or the existence of diverse data \cite{hendrycks2019augmix, zhang2022exact, zhoudomain} in the source domain to learn such robust feature representations. However, the training data in the AD task consists of only one class, and the data is monotonous. Consequently, it is difficult to adapt existing OOD generalization techniques to address the AD under distribution shift problem. Trivial adaption of the OOD generalization can fail to learn generalized normality representations, leading to many detection errors, \eg, normal samples with distribution shifts cannot be distinguished from anomalous samples and consequently they are detected as anomaly. As shown by the exemplar data in Fig. \ref{fig:intro}, normal samples in the in-distribution (ID) test data are very similar to the normal training data, and ID anomalies deviate largely from the normal data; however, due to the distribution shift, the normal samples in the OOD test data are substantially different from the ID normal data in terms of foreground and/or background features, and as a result, these normal samples can be falsely detected as anomaly. 

In this paper, we tackle the problem of anomaly detection under distribution shift. It is an \textit{OOD generalization} problem, aiming at learning generalized detection models to accurately detect normal and anomalous samples in test data with distribution shifts, while maintaining the effectiveness on in-distribution test data. This is different from the problem of \textit{OOD detection} \cite{
hendrycks2017baseline,hsu2020generalized,liu2020energy,ren2019likelihood,wang2022partial} that aims to equip supervised learning models with a capability of rejecting OOD/outlier samples as unknown samples for the sake of model deployment safety. This work makes three main contributions in addressing the OOD generalization problem in the AD task:
\begin{itemize}
\item We present an extensive study of the distribution shift problem in AD and establish large performance benchmarks under various distribution shifts using four widely-used datasets adapted from AD and OOD generalization tasks. Our empirical results further reveal that existing state-of-the-art (SOTA) AD and OOD generalization methods fail to work effectively in identifying anomalies under distribution shift.
\item We then propose a novel robust AD approach to diverse distribution shifts, namely \textit{generalized normality learning} (\textbf{GNL}). GNL minimizes the distribution gap between ID and OOD normal samples in both the training and inference stages in an unsupervised way. To this end, we introduce a normality-preserved loss function to learn distribution-invariant normality representations, which enables GNL to learn generalized semantics of the normal training data at different feature levels. GNL also utilizes a test time augmentation method to further reduce the the distribution gap during the inference stage.
\item Extensive experiments show that our approach GNL substantially outperforms state-of-the-art AD methods and OOD generalization methods by over 10\% in AUCROC on data with various distribution shifts, while maintaining the detection accuracy on the ID test data.
\end{itemize}
\section{Related Work}

\subsection{Anomaly Detection} 

\textbf{One-class Classification.}
Some early methods for anomaly detection include one-class support vector machine (OC-SVM) \cite{scholkopf2001estimating} and support vector data description (SVDD) \cite{tax2004support}. More recently, Deep SVDD \cite{ruff2018deep} uses a deep neural network to identify anomalies with a SVDD objective. A number of methods \cite{chen2022deep,goyal2020drocc,wu2019deep,sabokrou2020deep,yi2020patch} is then introduced to learn more effective deep one-class description.

\textbf{Reconstruction-based Methods.} One popular AD approach is to use autoencoder (AE) \cite{kingma2013auto}. AE-based anomaly detection learns normal patterns from a dataset to reconstruct new samples, assuming that anomalous samples have higher reconstruction errors due to distribution differences. There are many works following this direction and gaining good performance \cite{gong2019memorizing,hou2021divide,park2020learning,yan2021learning,zavrtanik2021reconstruction,pourreza2021g2,zaheer2020old}. 

\textbf{Self-supervised Learning Methods.} The use of data augmentation techniques is becoming increasingly prevalent in AD. One such strategy involves incorporating synthetic anomalies into datasets that are otherwise free of anomalies \cite{li2021cutpaste, yan2021learning, zavrtanik2021reconstruction}. 

\textbf{Knowledge Distillation.} Another popular line of research is knowledge distillation-based methods. A student-teacher framework with discriminative latent embeddings is introduced in \cite{bergmann2020uninformed}. Many improved versions for AD are then introduced \cite{salehi2021multiresolution,wang2021student,deng2022anomaly}. Anomaly Detection via Reverse Distillation (RD4AD) \cite{deng2022anomaly} is the latest one and gains SOTA performances on many datasets. 

All these methods are focused on AD with the same distribution in training and test data, which fail to work well on data with distribution shift.

\subsection{OOD Generalization} 

\textbf{Data Augmentation.} One popular approach for OOD generalization is based on data augmentation. Methods in this line involve generating new data samples from existing ones to increase the size and diversity of the training data. The model can then learn more about the underlying data distribution and become more robust to changes in the test data \cite{otalora2019staining, chen2020improving, zhang2020generalizing, sinha2017certifying, hendrycks2019augmix, zhang2022exact}.

\textbf{Unsupervised Learning.} By solving pretext tasks, a model can develop general features that are not specific to the target task. As a result, the model is less likely to be influenced by biases that are unique to a particular domain, which helps to avoid overfitting and increase generalization ability to different unseen data \cite{carlucci2019domain, ghifary2015domain, wang2020learning, maniyar2020zero, bucci2021self, albuquerque2020improving}.

Although these two types of methods are not designed for AD, they can be easily adapted for AD as they do not require class or domain labels during training. On the other hand, many existing OOD generalization methods, such as domain alignment \cite{muandet2013domain, hu2020domain, jin2020feature, mahajan2021domain, li2020domain, li2018domain, zhao2020domain}, meta-learning \cite{dou2019domain, li2021sequential, du2020learning, du2021metanorm, wang2020meta}, and disentangled representation learning \cite{li2017deeper, khosla2012undoing, chattopadhyay2020learning, piratla2020efficient, ilse2020diva, wang2020cross}, require class/domain-related supervision, which are inapplicable for the AD task. A similar issue exists for OOD generalization methods designed for multi-class problems \cite{du2020learning, huang2020self}. There are some cross-domain AD methods \cite{lu2020few, lv2021learning,ding2022catching}, but they require class labels in the ID data or few training samples from the target domain. By contrast, we focus on unsupervised AD and do not require any OOD data available during training. They focus on video data, while we focus on image data. Additionally, another related research line is on AD in situations involving a `near distribution' scenario \cite{mirzaei2022fake}, where anomalies are semantically similar to the normal distribution. Methods in this line can be more robust to distribution shift than general AD methods, but they do not tackle variations between the distributions of training and testing normal data.

\section{Problem Formulation and Challenges}

\subsection{Problem Formulation}
Let $\mathcal{X}_s$ and $\mathcal{X}_t$ denote the source (ID) and target (OOD) distributions, respectively, where $\mathcal{X}_s$ is used for both training and testing phase,  while $\mathcal{X}_t$ is only used for inference period. We assume that during training, only normal data from $\mathcal{X}_s$ is available, \ie, $\mathcal{D}_s = \{{x \in \mathcal{X}_s \mid y = 0}\}$, where $y \in \{0,1\}$ is the binary label indicating whether $x$ is a normal ($y=0$) or abnormal ($y=1$) sample. During testing, data can be normal or abnormal, and can be from either the source or target distribution, \ie, $\mathcal{D}_t = \{{ x \in \mathcal{X}_s \cup \mathcal{X}_t}\ \mid y = \{0,1\}\}$. The goal is then to develop an unsupervised anomaly detection model that can effectively handle distribution shift and accurately detect anomalies in $\mathcal{D}_t$. Specifically, we aim to learn a function $f:\mathcal{X}\rightarrow \mathbb{R}$ that assigns an anomaly score to each sample $x$ in a way such that $\forall x_i, x_j \in \mathcal{D}_t, \ f(x_i) < f(x_j) $ when $y_i=0$ and $y_j=1$.

\subsection{The Challenges} 
The current approaches in AD involve explicit fitting of the normal training data \cite{deng2022anomaly, salehi2021multiresolution, akcay2019ganomaly, schlegl2019f}. It can cause the model to learn irrelevant features that are not associated with the appearance of normal data, \eg, the model may mistake domain-specific background information as normal features, resulting in inaccurate anomaly detection when there are distribution shits presented. OOD generalization models are also significantly challenged by the studied setting.
This is mainly because the training data in the AD task consists of only one class and the data is monotonous, making it difficult to learn and identify patterns that distinguish normal and anomalous instances. Current OOD generalization approaches used in classification, detection, and segmentation need to take into account class labels, domain labels, or the diversity of samples in the training data \cite{du2020learning, huang2020self,wang2020cross, wang2020meta}, which often are not applicable to AD tasks. As a result, new methods are required that can effectively address the problem of AD under distribution shift.

\begin{figure}[t]
    \centering
    \includegraphics[width=0.5\textwidth]{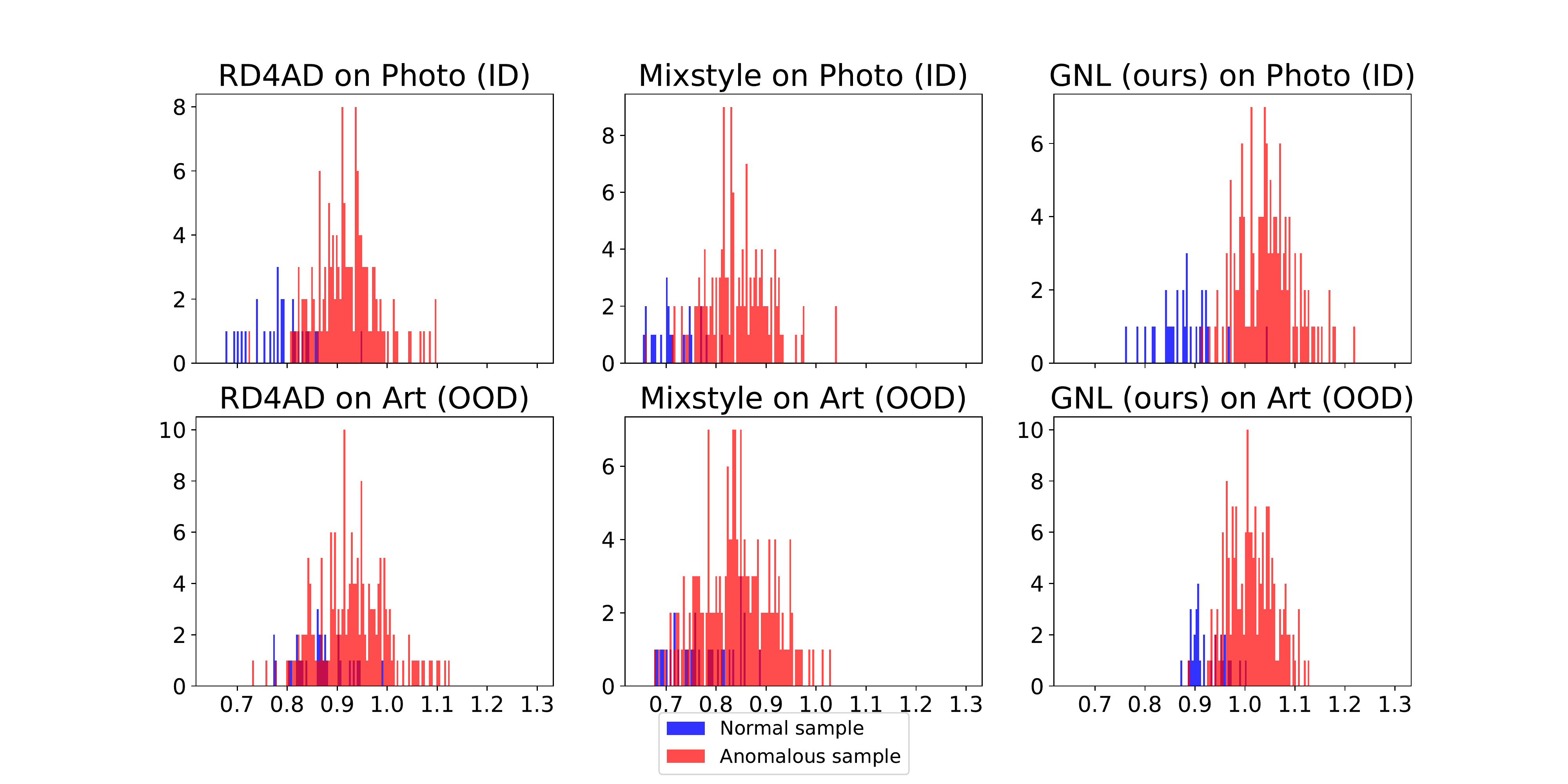}
    \caption{Anomaly scores of RD4AD \cite{deng2022anomaly}, Mixstyle \cite{zhoudomain} and our model GNL on PACS \cite{li2017deeper} when selecting `house' as the normal class and the remaining classes as anomaly classes.}
    \label{fig:challenge}
\end{figure}

Fig. \ref{fig:challenge} illustrates this issue, where models such as RD4AD \cite{deng2022anomaly} (a recent SOTA AD model) and Mixstyle \cite{zhoudomain} (an OOD generalization method that we use to combine with RD4AD) are seen to struggle with identifying normal samples in the presence of distribution shift, often misclassifying them as anomalous. The overlapping of histograms of the anomaly scores for the normal and abnormal samples indicates that these models have learned features that are not representative of normal data, which can be a major obstacle in detecting anomalies. One of the main reasons is that the background or style features w.r.t. a specific dataset can change due to different natural conditions. As a result, the model may mistake these changed features as anomalies, leading to normal samples in the shifted distribution being classified as anomalous with high anomaly scores. Furthermore, some abnormal samples in the OOD data may possess similar features to background or style features in the training data, leading to them being misclassified with low anomaly scores.

\section{Our Approach}
\begin{figure*}[h]
    \centering
    \includegraphics[width=\textwidth]{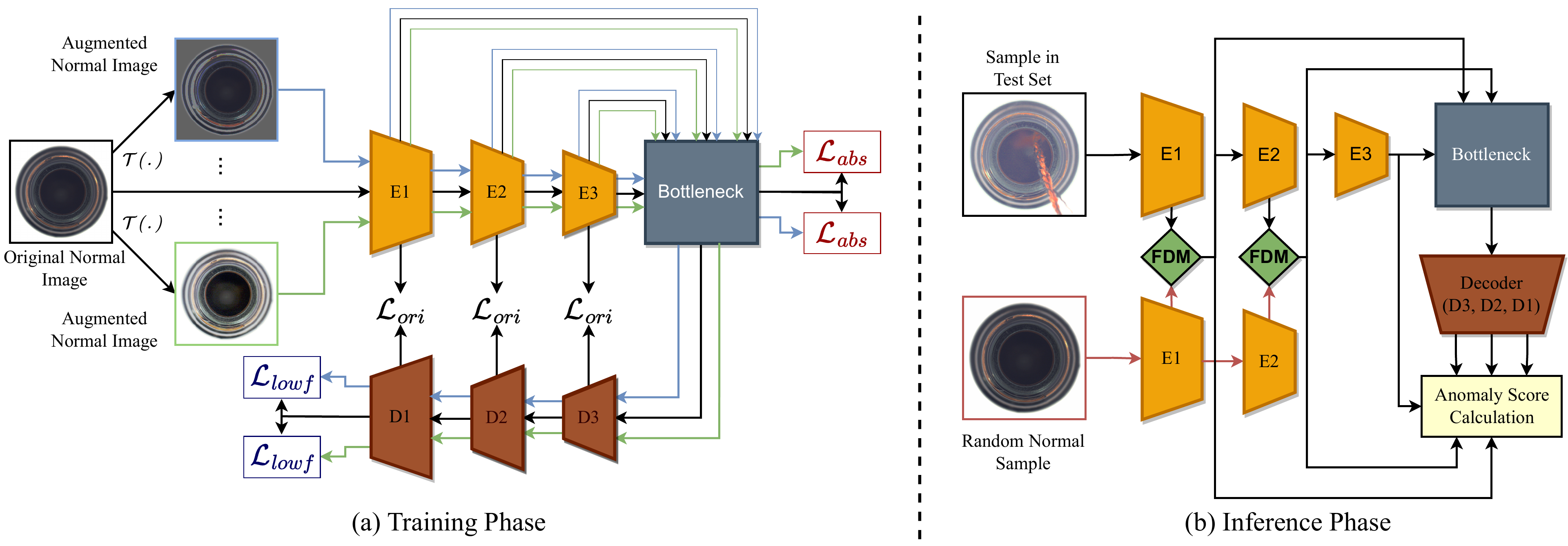}
    \caption{Overview of our approach. (a) Distribution-invariant normality learning in the training phase. (b) Test time augmentation with feature distribution matching in the inference phase.}
    \label{fig:proposedframework}
\end{figure*}

To address these challenges, we introduce a novel approach, namely generalized normality learning (GNL). GNL minimizes the distribution gap between ID and OOD normal samples in both the training and inference stages in an unsupervised way. To this end, we introduce a normality-preserved loss function to learn distribution-invariant normality representations, which enables GNL to learn generalized semantics of the normal training data at different feature levels. GNL further utilizes an AD-oriented test time data augmentation method based on feature distribution matching to improve the generalization performance. Fig. \ref{fig:proposedframework} describes the two main components of our approach: (a) distribution-invariant normality learning for training, and (b) test time augmentation methods. 
The two components complement to each other, meaning that the distribution-invariant normality learning process used during training can support the test time augmentation methods used during testing, and vice versa. 

\subsection{Distribution-invariant Normality Learning}

In order to improve the performance of model on OOD datasets while maintaining good performance on ID datasets, we aim to train a student model to be more robust to changes in the distribution of data, while still ensuring that the student overfits on the normal features. 
Fig. \ref{fig:proposedframework} (a) illustrates the training framework. 


Our method is built on top of the RD4AD model introduced by Deng et al. \cite{deng2022anomaly} that achieves state-of-the-art results on various datasets. The RD4AD framework includes three components: a fixed teacher encoder, a trainable one-class bottleneck embedding module, and a student decoder. When given an input sample, the teacher encoder extracts multi-scale representations, and the student decoder is trained to reconstruct the features from the bottleneck embedding. During testing, the teacher encoder can identify abnormal and OOD features in anomalous samples, but the student decoder fails to reconstruct these features. The model then considers anomalous representations that have low similarity as highly abnormal.

We propose to incorporate a similarity loss that quantifies the difference between the embedding features of the original samples and those of each transformed normal sample that represents a distinct style from the original data. Specifically, we enforce this loss at both the bottleneck layer and the final block of the decoder. To provide further clarity, we propose the inclusion of a loss term, denoted as $\mathcal {L}_{abs} $, which is integrated at the bottleneck layer of the encoder. Moreover, we also introduce another loss term, termed as $\mathcal {L}_{lowf}$, that is added at the final block of the student decoder architecture. Particularly, given a sample $x \in \mathcal{D}_s$, we
first apply an augmentation function $\mathcal{T(.)}$ on it, and let $x'_k =  \mathcal{T}(x)$ where $k \in [1,N]$ with $N$ is the number of augmented normal samples generated by data augmentation,
and $\phi$ be the mapping that projects the raw image $I$ into the embedding space at the bottleneck layer, then we define $\mathcal {L}_{abs} $ as:
\begin{equation}
 \mathcal{L_{\text{abs}}} = \sum _{k=1}^{N}\frac {1}{N} \bigg \{ \mathcal {L}_{sim}(\phi(x),\phi(x'_k))\bigg \} 
 \label {eq:lossabtract}, 
\end{equation}
where $\mathcal {L}_{sim}(.,.)$ is a cosine similarity-based loss function. 

Let $\omega$ be a reconstruction function from the abstract features to the low-level features at the final block of the decoder, then we further define $\mathcal {L}_{lowf} $ as:
\begin{equation}
 \mathcal {L}_{lowf}  = \sum _{k=1}^{N}\frac {1}{N} \bigg \{ \mathcal {L}_{sim}(\omega(\phi(x)),\omega(\phi(x'_k)))\bigg \}. 
 \label{eq:losslowfeature}
\end{equation}

We combine these loss functions to introduce the distribution-invariant, normality-preserved loss function:
\begin{equation}
 \mathcal {L} = \lambda _{ori}*\mathcal {L}_{ori} + \lambda _{abs} * \mathcal {L}_{abs} + \lambda _{lowf} * \mathcal {L}_{lowf} 
  \label{eq:sumloss}, 
\end{equation}
where $\mathcal{L}_{ori}$ is the original loss of RD4AD, and $\lambda _{ori}$, $\lambda _{abs}$, and $\lambda _{lowf}$ are hyperparameters that determine how much weight should be given to each type of loss function.

We adopt AugMix \cite{hendrycks2019augmix} as the data augmentation method. Still, we remove the augmentation types that have the potential to generate anomalies, \eg, `shear\_x', `shear\_y', `translate\_x', and `translate\_y', to ensure that all generated data are normal samples.

Intuitively, the last block of decoder is responsible for reconstructing simple and low-level features, such as edges, corners, and blobs, while the bottleneck layer is responsible for extracting more complex and high-level features. At the bottleneck layer, the abstracted information of the same images from different synthesized methods must be the same, while retaining enough information for reconstruction in the decoder. Therefore, by minimizing the loss function in Eq. \ref{eq:sumloss}, GNL learns features from both low-layer CNNs and high-level CNNs respectively to be the same from different distributions generated from a single sample. 

\subsection{Test Time Augmentation for Anomaly Detection under Distribution Shift}

The goal of this component is to address the problem of a mismatch between the distribution of data during testing. To accomplish this, we propose injecting training distribution into the inference samples by using Feature Distribution Matching (FDM) at multi-level layers of the teacher encoder in the inference phase. The proposed testing framework is demonstrated in Fig. \ref{fig:proposedframework} (b). Our test time augmentation is applied at the first two residual blocks of the teacher encoder. The inference process from the third residual block onwards, as well as the calculation of the anomaly score, follow the original RD4AD framework without any modifications.

FDM is a group of techniques that aims to reduce the distribution mismatch or discrepancy of data from two different domains. Some previous studies focused on FDM assume that the input features follow a Gaussian distribution \cite{huang2017arbitrary, lu2019closed, mroueh2020wasserstein}. More recently, Zhang et al. \cite{zhang2022exact} introduced a more accurate approach, known as Exact Feature Distribution Matching (EFDM). EFDM precisely matches empirical Cumulative Distribution Functions of image features,
resulting in exact feature distribution alignment (as the sample size tends to infinity) and accurate matching of statistical properties like mean, standard deviation, and high-order statistics. Basically, all these FDM techniques are applicable to our proposed framework. Noted that FDM have been used for OOD Generalization, \eg, in Mixstyle \cite{zhoudomain} and EDFMix \cite{zhang2022exact}, but they are used during training with the goal of creating new distribution samples by mixing the subdomain of the samples available in the training set, while we adopt FDM as a component in the inference stage with a different objective.

Specifically, given a test sample $p \in \mathcal{D}_t$, we randomly select a training normal sample $ q \in \mathcal{D}_s$. These two samples are then fed into the teacher encoder. Let $\mathcal P^m$ and $\mathcal Q^m$ be the embedded features of $p$ and $q$ at the residual encoding block $E^m$, respectively, then the testing process is performed as follows:
\begin{equation}
\begin{cases}
\mathcal P^{m+1} = \text{FDM}(E^{m+1}(\mathcal P^{m}), \mathcal Q^{m+1},\alpha) \\
\mathcal Q^{m+1} = {E}^{m+1}(\mathcal Q^{m})\\
\mathcal P^{0} = p, \mathcal Q^{0} = q,\\

\end{cases}
\label{equa:test}
\end{equation}
where $ m \in \{0,1\}$ and $\alpha$ is a hyperparameter balancing the severity for mixing the style between the inference sample and the selected normal sample. The processed embedded features $\mathcal P^{1} \text{ and } \mathcal P^{2}$ are then input into the bottleneck layer and participate in the calculation of anomaly scores following the inference process of the original RD4AD.

For the $\text{FDM}()$ function above, EFDM\cite{zhang2022exact}, which is the SOTA of FDM, is adopted to our method as follow:
\begin{equation}
\textrm {FDM}(\mathcal{C} ,\mathcal{V} , \alpha ): \mathcal{C} _{\tau _i} = (1-\alpha) {\mathcal{C} _{\tau_i}} + \alpha  {\mathcal{V} _{\kappa _i}} 
  \label {eq:sumloss}, 
\end{equation}
where ${\{\mathcal{C} _{\tau _i}\}}^n_{i=1}$ and ${\{\mathcal{V} _{\kappa _i}\}}^n_{i=1}$ are sorted values of embedded feature $\mathcal{C}$ and $\mathcal{V}$ in ascending order.  Here, $n$ represents the number of elements in vector $\mathcal{C}$ and $\mathcal{V}$. Note that $\mathcal{C}$ is the embedded feature of the test sample $p$, which plays the role of carrying the appearance information. $\mathcal{V}$ is the embedded feature of a normal sample $q$ randomly sampled from the training data, carrying the style information.

In essence, the sample $q$ plays a role in conveying distribution information pertaining to the training data. The selection of a random sample is due to the monotonous nature of the data during training, as any sample in the training set is capable of carrying distribution information that represents the training data. Thus, It helps avoid a process for careful sample selection that is often computationally expensive.

By utilizing FDM, our proposed testing process minimizes the disparity between the feature distribution of the inference sample and the feature distribution of normal samples in the training data, in cases where inference samples come from OOD sets. Furthermore, FDM ensures that the feature distribution remains nearly unchanged if inference samples come from ID sets, since the distribution of the test sample is aligned with its own distribution. Therefore, our testing approach can improve performance on OOD data without sacrificing performance on ID data.

\section{Experiments}
\subsection{Datasets}
We adapt four datasets from both AD and OOD generalization as the dataset benchmarks for the studied task.

\textbf{Anomaly Detection.}
\textbf{MVTec} \cite{bergmann2019mvtec} is a widely-used AD benchmark, which comprises 15 data subsets for industrial defect inspection,
including 5 subsets on texture anomalies and 10 subsets on object anomalies. The training dataset consists of 3,629 images in total, all of which are normal images. In contrast, the test dataset contains a total of 1,725 images, comprising both defective and non-defective instances. \textbf{CIFAR-10 \cite{krizhevsky2009cifar}} serves as a one-class classification benchmark, featuring 50,000 training and 10,000 test images across 10 equally-sized categories representing diverse natural entities. In order to generate OOD datasets for MVTec and CIFAR-10, we apply 4 types of visual corruptions~\cite{hendrycks2019robustness} to MVTec and CIFAR-10: Brightness, Contrast, Defocus Blur, and Gaussian noise. The severity for each type of corruption is set to 3 on MVTec and 5 on CIFAR-10 for obtaining the out-of-distribution data. 

\textbf{OOD Generalization.} 
Two popular OOD benchmarks, \textbf{MNIST-M} \cite{ganin2015unsupervised} and \textbf{PACS} \cite{li2017deeper}, are taken in our experiments.
In particular, the primary MNIST \cite{lecun1998gradient} is used as the ID data on which the models are trained on, while MNIST-M is used as the OOD set. MNIST and MNIST-M datasets share 10 classes, which correspond to the digits 0 through 9. While MNIST encompasses 70,000 grayscale images of handwritten digits, MNIST-M contains 68,000 OOD images that are synthesized by superimposing random colored patches on the original images from MNIST. PACS is another widely used OOD dataset consisting of 9,991 images, which are shared by seven classes and four domains, namely Art, Cartoon, Photo, and Sketch. We select the images in Photo as the ID data, with the images in Art, Cartoon, and Sketch as the OOD data. 
The commonly used one-versus-all protocol \cite{perera2019ocgan} is used to convert the these two datasets into AD datasets with distribution shift, in which samples of one class are used as normal, with the rest of classes as anomaly classes. Furthermore, we perform a multi-class setting on the MNIST/MNIST-M dataset, labeling samples from even-numbered classes as normal, while those from odd-numbered classes are identified as anomalies.

During training, we only use images in the ID dataset, \ie, assuming the OOD data is not available during training. During inference, test sets of both ID and OOD are used.

\subsection{\textbf{Baselines}}
We conduct a series of experimental evaluations on 4 prominent anomaly detection methods, namely Deep SVDD \cite{ruff2018deep}, f-AnoGAN \cite{schlegl2019f}, KDAD \cite{salehi2021multiresolution}, and RD4AD \cite{deng2022anomaly}. These methods stand for popular AD methods and recent state-of-the-art (SOTA) AD models.
To evaluate the efficacy of OOD generalization techniques in anomaly detection, we adapt a suite of cutting-edge OOD methods by combining them with the recently proposed RD4AD model, which boosts SOTA performance on multiple datasets. Four different methods are used, including three data augmentation-based methods Augmix \cite{hendrycks2019augmix}, Mixstyle \cite{zhoudomain}, and EFDM \cite{zhang2022exact}, and one self-supervised method Jigsaw \cite{carlucci2019domain}. 

\subsection{Implementation Details}
Our proposed method GNL is implemented on top of the RD4AD framework. Therefore, we maintain the settings recommended by RD4AD, such as the image size, the optimization method, the way of calculating anomaly score, and other relevant parameters. The details can be found in Appendix. Regarding the specific parameters for our model GNL, we choose $N=2$ for the number of augmented normal samples generated by data augmentation. We set $\lambda_{ori}=0.9, \lambda_{abs}=0.05 \text{ and } \lambda_{lowf}=0.05$ by default for the distribution-invariant, normality-preserved loss function. During the inference phase, we opt for $\alpha=0.5$ to control the degree of style blending for MVTec, PACS, and CIFAR-10 datasets, while setting $\alpha=0.9$ for MNIST/MNIST-M, effectively managing the mixing dynamics. We choose EFDM \cite{zhang2022exact} as the FDM technique since it is the latest and shows SOTA performance. 

For the AD baselines, we use the official implementation published by the authors of those baselines. 
However, since the original baselines did not include experiments on the PACS dataset, we use the hyperparameters from MVTec experiments to conduct experiments on the PACS dataset corresponding to each baseline. 

For the OOD generalization baselines, we use Augmix with an online augmentation severity of 3. We use all the data augmentation types included in Augmix for MNIST and PACS. However, for the MVTec and CIFAR-10 dataset, we exclude two types of augmentation that overlap with two types of corruptions during testing: Brightness and Contrast. With Mixstyle and EFDM, which are two data augmentation methods at the feature level (rather than at the image level like Augmix), we apply Mixstyle and EFDM to the encoders in the first two network layer according to the settings in RD4AD. As for Jigsaw, we fit the Jigsaw task into the Bottleneck component in RD4AD. All hyperparameters of training are preserved when applying OOD generalization baselines into RD4AD.

Following previous studies \cite{ruff2018deep, schlegl2019f, salehi2021multiresolution, deng2022anomaly,ding2022catching}, we evaluate the performance of our anomaly detection methods using a metric called the Area Under the ROC Curve (AUROC). This metric is commonly used to assess how well a given method is able to distinguish between normal and anomalous data points. The results are averaged over three independent runs.

\subsection{Comparison Results}





The performance of our model GNL and the baselines on MVTec, CIFAR-10, MNIST, and PACS are shown in Tables \ref{tab:mvtec}, \ref{cifar10}, \ref{tab:mnist} and \ref{tab:pacs}, respectively. Note that due to space limitations, the performances in all four tables are the average results of the classes per dataset. Detailed results are presented in Appendix. Overall, GNL can significantly outperforms SOTA AD models and OOD generalization methods in detecting anomalies on the OOD test data, while at the same time maintaining the detection accuracy on the ID data. Below we discuss the results in detail.

\subsubsection{Performance of AD Methods}
In general, we observe a significant drop in the AUC scores of all AD methods, Deep SVDD, f-AnoGAN, KDAD and RD4AD, on the OOD data across all four datasets used. This indicates that their performance is severely affected by the distribution shift. In particular, the performance of all AD models is promising on the MNIST set. However, this performance is reduced by about 30-40\% when the models are tested on the MNIST-M set, which contains variations that are not present in the original MNIST set. Similar trends are observed in the PACS dataset, where the models' performance is also significantly affected by the distribution shifts in the OOD data. The models perform well on the Photo data, which is the ID data, but their performance drops significantly on the three OOD datasets, Art, Cartoon and Sketch. On MVTec and CIFAR-10, the performance still drops but is less severe than on the other two sets.

\begin{table}[h]
\centering
\scalebox{0.8}{
\begin{tabular}{l|c|cccc}
\hline
& \multicolumn{1}{c}{ID} & \multicolumn{4}{|c}{OOD} \\
\cline{2-5}
\hline
Method & MVTec & Brightness & Contrast & Blur & Noise \\
\hline
Deep SVDD & 69.98 & 55.18 & 50.07 & 68.82 & 59.11 \\
f-AnoGAN & 75.65 & 48.36 & 49.29 & 37.98 & 39.10\\
KDAD & 85.50 & 83.81 & 64.03 & 84.17 & 82.04 \\
RD4AD & \textbf{98.64} & 96.50 & 94.12 & \textbf{98.9} & 90.14 \\
\hline
Augmix & 96.29 & 95.10 & 94.51 & 95.39 & 90.99 \\
Mixstyle & 98.58 & 96.60 & 94.45 & 98.27 & 88.92 \\
EFDM & 98.64 & 96.78 & 94.77 & 98.25 & 89.29 \\
Augmix+Mixstyle & 96.78 & 96.86 & 94.57 & 98.73 & 90.12 \\
Augmix+EFDM & 97.04 & 96.83 & 95.21 & 98.11 & 90.18  \\
Jigsaw & 73.97 & 73.36 & 67.88 & 73.88 & 72.60 \\
\hline
GNL (Ours) & 97.99 & \textbf{97.43} & \textbf{97.46} & 97.77 & \textbf{94.10} \\
\hline
\end{tabular}
}

\caption{AUROC (\%) results on MVTec and its four corruptions. The best performance is \textbf{boldfaced}.}
\label{tab:mvtec}
\end{table}

\begin{table}[h]
\centering
\scalebox{0.8}{
\begin{tabular}{c|c|cccc}
\hline
& \multicolumn{1}{c}{ID} & \multicolumn{4}{|c}{OOD} \\ \hline
Method &CIFAR & Brightness & Contrast & Blur & Noise  \\
\hline
Deep SVDD & 64.62 & 59.13 & 55.94  & 62.13 & 54.46 \\
f-AnoGAN & 70.25 & 54.62 & 57.23  & 60.74 & 51.76 \\
KDAD & 84.21 & 75.91 & 64.37  & 63.49 & 56.87 \\
RD4AD & \textbf{84.62} & 75.89 & 65.34  & 66.67 & 58.82 \\ \hline
Augmix & 82.83 & 74.15 & 62.48  & \textbf{66.92} & 57.36\\
Mixstyle & 83.68 & 76.07 & 63.87  & 65.74 & 57.74\\
EFDM & 83.92 & 76.19 & 63.92  & 64.81 & 57.63\\
Augmix+Mixstyle & 83.87 & 76.02 & 65.55 & 63.89 & 58.04 \\
Augmix+EFDM & 82.96 & 75.73 & 64.39 & 63.83 & 57.14 \\
Jigsaw & 71.29 & 66.86 & 61.45 & 60.12 & 55.29 \\\hline
Ours & 82.29 & \textbf{77.94} & \textbf{66.13}  & 64.04 & \textbf{61.51} \\
\hline
\end{tabular}}
\caption{AUROC (\%) results on CIFAR-10 and its four corruptions.}
\label{cifar10}
\end{table}

\begin{table}[h]
\centering
\scalebox{0.9}{
\begin{tabular}{c|cc|cc}
\hline
& \multicolumn{2}{c}{One-vs-All} & \multicolumn{2}{|c}{Multi-class} \\
\hline
Method  & ID & OOD & ID & OOD\\
\hline
Deep SVDD & 97.73 & 49.92 & 86.94 & 51.19 \\
f-AnoGAN & 97.52 & 52.72 & 88.45 & 51.85\\
KDAD & 98.87 & 54.87 & \textbf{90.43} & 52.84\\
RD4AD & \textbf{98.89} & 58.09 & 88.70 & 51.74\\
\hline
Augmix & 98.26 & 59.61 & 88.76 & 52.19\\
Mixstyle & 98.84 & 57.22 & 87.36 & 52.13\\ 
EFDM & 98.62 & 57.23 & 87.78 & 52.36\\
Augmix+Mixstyle & 98.12 & 58.89 & 89.23 & 52.45  \\
Augmix+EFDM & 98.24 & 58.91 & 90.04 & 52.64\\
Jigsaw & 98.90 & 58.51 & 87.29 & 52.87\\
\hline
GNL (Ours) & 96.91 & \textbf{70.87} & 88.59 & \textbf{58.50}\\
\hline
\end{tabular}
}
\caption{AUROC results (\%)  on in-distribution (MNIST) and out-of-distribution (MNIST-M) datasets for one-vs-all and multi-class settings.}
\label{tab:mnist}
\end{table}

\begin{table}[h]
\centering
\scalebox{0.9}{
\begin{tabular}{l|c|ccc}
\hline
& \multicolumn{1}{c}{ID} & \multicolumn{3}{|c}{OOD}  \\
\cline{2-5}
\hline
Method & Photo & Art & Cartoon & Sketch  \\ \hline
Deep SVDD   & 40.87&	53.42 &	41.23 & 	39.48 \\
f-AnoGAN &  61.34 &  50.15 &  52.42 &  63.77   \\
KDAD &  \textbf{88.17} &  62.86 &  62.64 &  51.40 \\
RD4AD &  81.49 &  61.07 &  60.34 &  55.06  \\
\hline
Augmix &  76.35 &  60.50 &  58.96 &  57.86  \\
Mixstyle &  78.23 &  60.93 &  60.93 &  54.89 \\
EFDM &  78.47 &  60.55 &  62.15 &  55.63  \\
Augmix+Mixstyle & 76.12 & 60.16 &  61.29 & 55.76\\
Augmix+EFDM & 77.28 & 60.93 &  63.18 & 56.67\\
Jigsaw &  62.19 &  52.55 &  53.83 &  62.15  \\
\hline
GNL (Ours) &  87.67 &  \textbf{65.62} &  \textbf{67.96} &  \textbf{62.39}  \\ 
\hline
\end{tabular}
}
\caption{AUROC results (\%) on in-distribution (Photo) and out-of-distribution (Art, Cartoon, Sketch) datasets.}
\label{tab:pacs}
\end{table}


\subsubsection{Performance of Combined OOD Generalization and AD Methods}
Our results in Tables \ref{tab:mvtec}, \ref{cifar10}, \ref{tab:mnist} and \ref{tab:pacs} indicate that the detection performance cannot be significantly improved by combining different OOD generalization techniques with the recent SOTA AD model RD4AD on the four datasets. This lack of improvement can be attributed to the fact that these OOD methods attempt to increase the diversity of data by enriching the available data based on its own distribution. However, because the training data in AD is typically monotonous and unimodal, these OOD methods often fail to generate data samples that significantly deviate from the original data distribution. As a result, the added diversity of generated data is not sufficient to significantly improve the performance of AD models.

Moreover, these OOD techniques also have a tendency to generate undesired anomaly data, which is akin to injecting noise into the training data, thereby reducing the performance of AD models on the in-distribution dataset.

\subsubsection{Performance of Our Method GNL}
 
On the MVTec AD dataset in Table \ref{tab:mvtec}, our method shows remarkable improvement in performance on the OOD dataset, while maintaining the performance on the ID data. In fact, our method achieves a highly comparable AUROC score of 0.9799 on the original MVTec ID data, while also obtaining an impressive AUROC score of 0.9743 on the Brightness, 0.9746 on Contrast, 0.9777 on the Defocus\_blur dataset, and 0.9410 on Gaussian Noise, which are significant improvements over the other methods. These results demonstrate the robustness and effectiveness of our GNL model to diverse distribution shifts.

Our experimental findings on the CIFAR-10 dataset exhibit a close resemblance to the outcomes observed on the MVTec dataset, as depicted in Table \ref{cifar10}. Our approach attains a competitive AUROC score of 0.8229 on the native MVTec ID data. Notably, our method yields enhanced AUROC scores of 0.7794 for Brightness, 0.6613 for Contrast, 0.6404 for Defocus\_blur, and 0.6151 for Gaussian Noise datasets, showcasing often large enhancements compared to other method, especially on the Contrast and Noise cases.

On the MNIST/MNIST-M dataset in Table \ref{tab:mnist}, GNL consistently and significantly outperforms all other methods on the OOD data MNIST-M, increasing by at least 10 AUROC scores. Compared to the best performer -- RD4AD -- on the ID dataset that obtains an AUROC score of 0.9889, GNL exhibits a small decline and obtains an AUROC score of 0.9691. However, a significant improvement in performance is observed on the MNIST-M dataset, with an AUROC score of 0.7087 compared to 0.5809 for RD4AD. Regarding the multi-class setting, the results indicate its increased challenge compared to one-vs-all setting. Our model still maintains superior performance on OOD data while also excelling on ID data.

Similarly, GNL achieves consistently more superior AUROC performance on all four OOD datasets of PACS in Table \ref{tab:pacs}. In particular, GNL obtains AUROC scores of 0.6562, 0.6796, and 0.6239 for the Art, Cartoon, and Sketch datasets, respectively, increasing by at least 5\% on the Art and Cartoon datasets over the competing models. Compared to RD4AD, our method not only largely improves the OOD performance, but also enhances its performance on the ID data, the Photo data. This is because the Photo data contains multiple sub-domains, and RD4AD can be susceptible to overfitting on a specific sub-domain in the training data. By contrast, our method helps to mitigate this issue by learning more generalized normality representations, which improves performance across all sub-domains within the Photo data. The performance of GNL on the ID data is also highly comparable to the best performer KDAD, 0.8767 vs. 0.8817, whereas GNL outperforms KDAD on the three OOD datasets by about 3\%-10\% in AUROC.


\subsection{Robustness to Various Distribution Shift Levels}

\begin{figure}[h]
    \centering
    \includegraphics[width=0.49\textwidth]{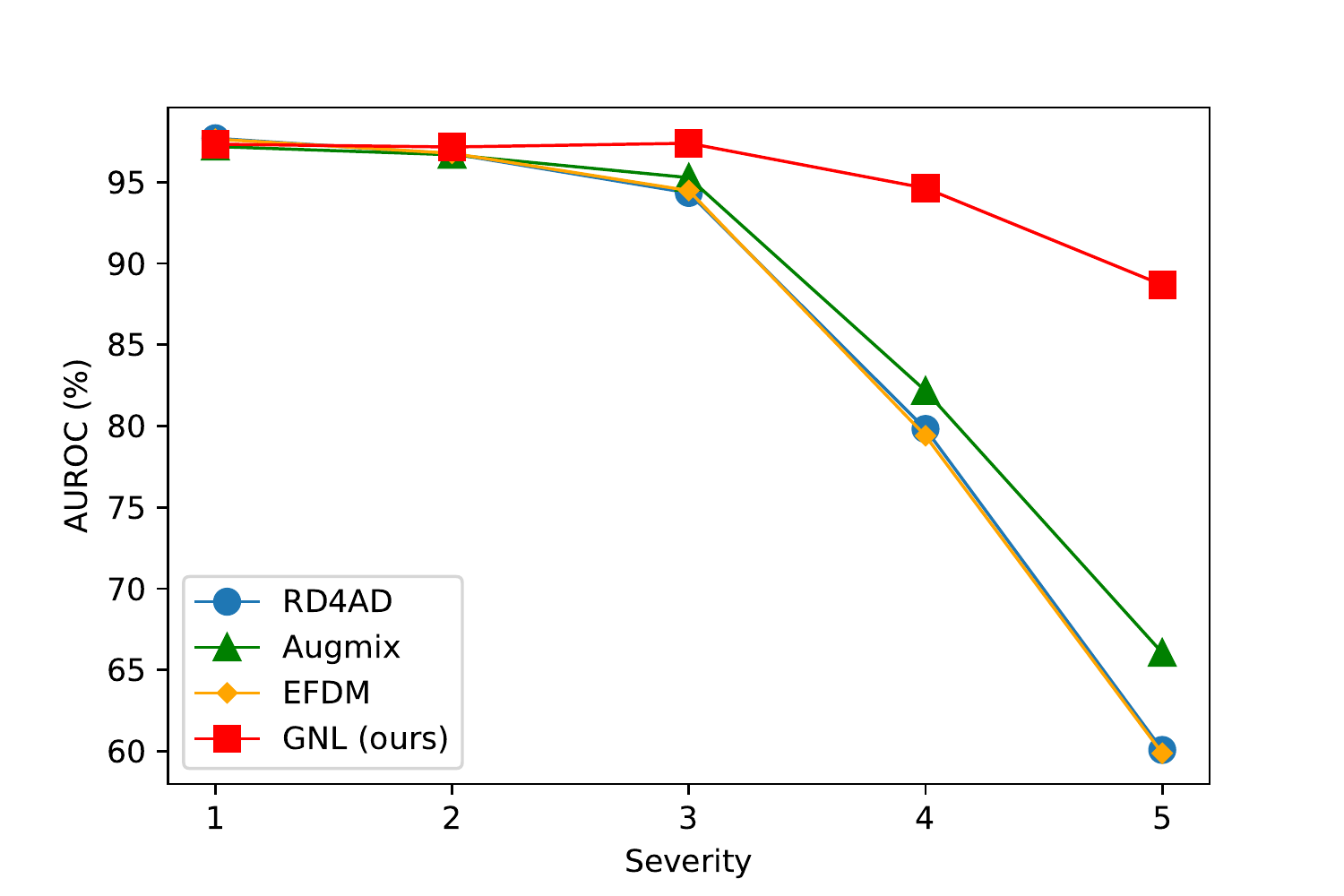}
    \caption{AUROC results on MVTec with varying severity of the `Contrast’ corruption.}
    \label{fig:changingseverity}
\end{figure}

Fig. \ref{fig:changingseverity} presents the results of the robustness of GNL to varying levels of distribution shift, using the best competing methods RD4AD, Augmix and EFDM as baselines. The experiments are done on MVTec with increasing levels of `Contrast' corruption. Notably, the performance of the baselines exhibits a significant decline as the severity of corruption amplifies.
The reason behind this phenomenon is intuitive as increased corruption severity introduces more substantial distribution variance, making it arduous for the models to discern between anomalous and normal samples. Our proposed method, on the other hand, demonstrates remarkable stability in performance across multiple levels of distribution shift. Our method maintains stable performance when the severity is between 1 and 3, and reduces to an AUROC of about 0.90 when the severity is 4 and 5, decreasing about 5\% AUROC vs. about 30\%-35\% decrease in the competing methods. These results indicate 
strong robustness of GNL to heavy distribution shifts.

\subsection{Ablation Study}

We examine the importance of two main components: Distribution-invariant Normality Learning (DINL) using $\mathcal{L}_{abs}$ and $\mathcal{L}_{abs}$ individually or simultaneously (in addition to $\mathcal{L}_{ori}$),  and AD-oriented Test Time Augmentation (ATTA) on the PACS dataset, with RD4AD as the baseline. The results are reported in Table \ref{tab:ablation}. The experiment results show that $\mathcal{L}_{abs} $ and $\mathcal{L}_{lowf} $ positively contribute to the superior performance of DINL from low-level and high-level features respectively; and they can complement each other when combining them in DINL. Looking more broadly, two main components, DINL and ATTA ,also positively contribute to the superior performance of GNL. In particular, the experimental results show that if only the test time augmentation is applied, we gain about 2\% AUROC improvement over the baseline on the OOD datasets, but it leads to a slight performance decrease on the ID data. 
When DINL is applied, it results in substantial improvement across both ID and OOD datasets, having 4\%-7\% AUROC improvement. When both are applied, we obtain the best performance, resulting in further substantial AUROC improvement. This indicates that both components, one reducing the distribution gap during training and another reducing the gap during inference, can well complement each other.

\begin{table}[h]
\centering
\scalebox{0.9}{
\begin{tabular}{c|c|ccc}
\hline
& \multicolumn{1}{c}{ID} & \multicolumn{3}{|c}{OOD} \\
\cline{1-5}

 Method & Photo & Art & Cartoon & Sketch \\
\hline
Baseline &  81.49 &  61.07 &  60.34 &  55.06 \\
$\mathcal {L}_{abs} $ only  & 82.02 & 60.59 & 63.93 & 56.81 \\
$\mathcal {L}_{lowf} $ only & 82.90 & 61.27 & 62.25 & 55.52 \\
DINL &  85.71 &  62.34 &  65.63 &  57.12 \\
ATTA &  81.05 &  64.36 &  62.04 &  57.04 \\
DINL+ATTA &  \textbf{87.67} &  \textbf{65.62} &  \textbf{67.96} &  \textbf{62.39} \\
\hline
\end{tabular}
}
\caption{AUROC results (\%) of ablation study.}
\label{tab:ablation}
\end{table}

\begin{figure}[h]
    \centering
    \includegraphics[width=0.49\textwidth]{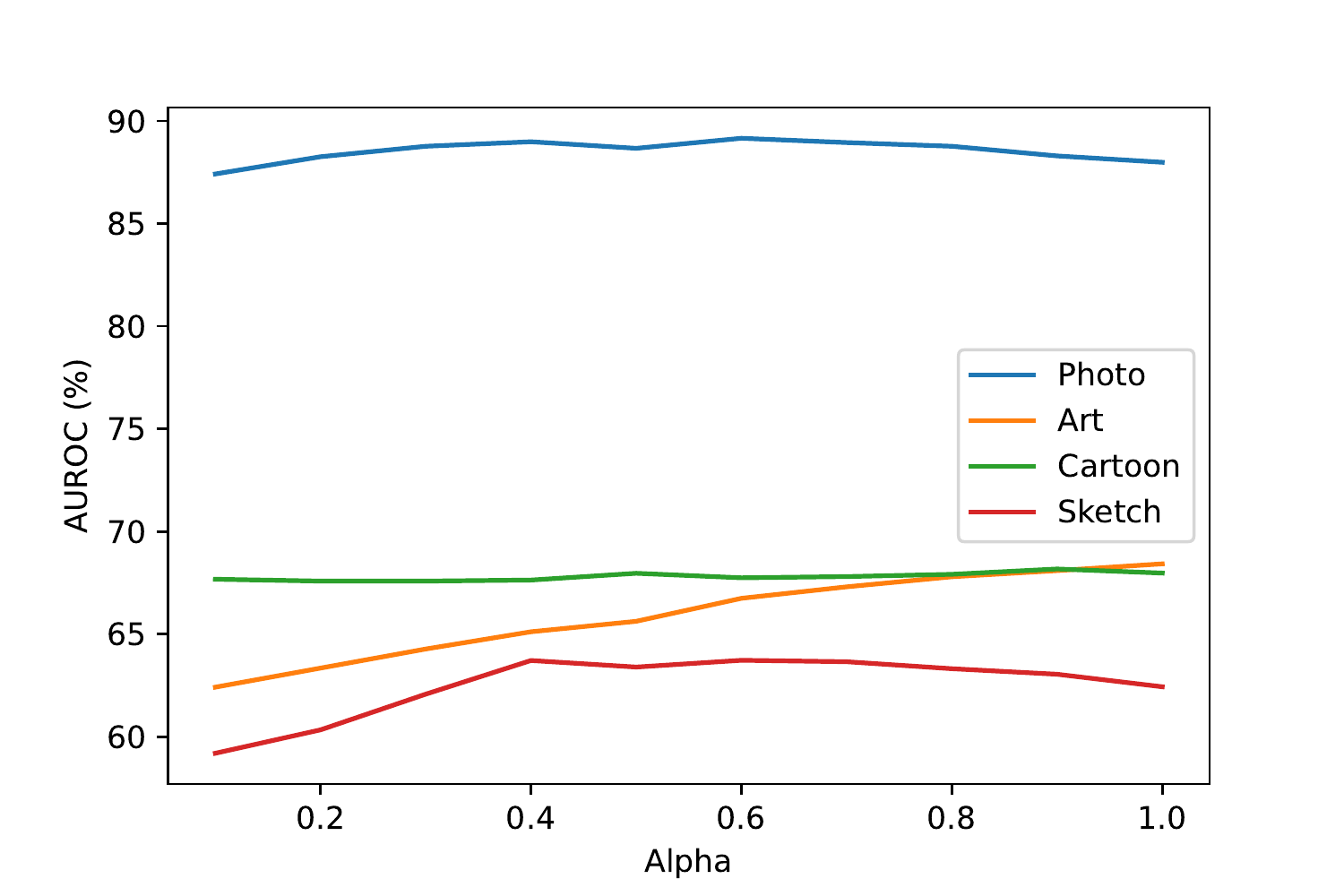}
    \caption{AUROC results using varying $\alpha$.
    The smaller the $\alpha$ value, the lower the severity of style transfer.}
    \label{fig:alpha}
\end{figure}

\subsection{Hyperparameter Analysis}
Fig. \ref{fig:alpha} depicts how the performance of our model GNL changes with varying $\alpha$, which is a hyperparameter in ATTA. 
The results suggest that the effectiveness of our model remains consistent across different $\alpha$ values on the Photo and Cartoon datasets. In contrast, the model's ability to detect anomalies on the Art data appears to improve as $\alpha$ increases. However, for the Sketch data, the model's performance reaches its maximum at $\alpha = 0.4$ and slightly decreases as $\alpha$ increases further. Overall, a medium value, \eg, $\alpha=0.5$, is generally recommended in practice.

We evaluate the hyperparameter sensitivity of our key component DINL using four settings of the three $\lambda$ hyperparameters: $\lambda _{ori}$, $ \lambda _{abs}$, and $\lambda _{lowf}$, with their sum set to one to ease the analysis. $ \lambda _{abs}=\lambda _{lowf}$ is used as the features learned by them are considered equally important for the task. The results on PACS are shown in Table \ref{tb2}. DINL shows good robustness across different hyperparameter ratios in the three losses. 

\begin{table}[h]
\centering
\scalebox{0.8}{
\begin{tabular}{c|c|ccc}
\hline
$\lambda _{ori}; \lambda _{abs}; \lambda _{lowf}$   &Photo (ID) & Art & Cartoon & Sketch \\
\hline
0.95, 0.025, 0.025 & 85.90 & 62.71 & 64.28 & 59.97 \\
0.90, 0.050, 0.050 & 85.71 & 62.34 & 65.63 & 57.12 \\
0.85, 0.075, 0.075 & 85.60 & 63.23 & 65.03 & 58.57 \\
0.80, 0.100, 0.100 & 84.89 & 61.45 & 65.12 & 56.84 \\
\hline
\end{tabular}}
\caption{AUROC using various $\lambda$ settings.}
\label{tb2}
\end{table}

\subsection{Time and Space Efficiency}
For space complexity, our method improves the training objective and the inference of RD4AD without altering its architecture, thereby avoiding any increase in the number of parameters. 

As shown in Table \ref{time}, in terms of time efficiency, our method's training duration is slightly longer than RD4AD, but this additional time yields substantial performance improvements. As for the inference, our approach remains reasonably responsive.

\begin{table}[h]
\centering
\scalebox{0.8}{
\begin{tabular}{ccc}
\hline
Model & Training (per epoch) & Inference (per image) \\
\hline
RD4AD & 2.5726  & 0.0282 \\
Ours & 7.2557 & 0.0356 \\
\hline
\end{tabular}}
\caption{Runtime (s) on the `Dog' dataset of PACS using one RTX 3090 24GB GPU.}\label{time}
\end{table}

\section{Conclusion}
In this work we propose a novel approach, namely GNL, to addressing the problem of anomaly detection in the presence of distribution shifts. GNL improves the generalization of the detection model by reducing the distribution gap between ID and OOD normal data in both training and inference stages. We also present comprehensive performance benchmarks and reveal that combined AD and OOD generalization methods do not work well for this task. Our approach is specifically designed for the OOD generalization in the AD task and shows significant improvement over the competing baselines. As shown in our results, our approach GNL is also robust to heavy distribution shifts.
Overall, our approach represents an important contribution to unsupervised anomaly detection, as it addresses a more realistic problem that has not been adequately studied before.



{\small
\bibliographystyle{ieee_fullname}
\bibliography{refs}
}
\appendix

\section{Full results on MNIST/MNIST-M, PACS and MVTec}
The full experimental results on the MNIST/MNIST-M, PACS and MVTec datasets are presented in Tables 1, 2, and 3, respectively. These tables provide a detailed illustration of the performance of the considered methods and the proposed method GNL. It is observed that GNL consistently outperforms the compared methods on many subsets of these datasets, indicating the effectiveness and robustness of GNL.

\section{Implementation details}
\subsection{Training}
With MVTec, all images in MVTec are resized to 256x256. We take ResNet50 \cite{he2016deep} as the backbone of the teacher encoder. The hyperparameters for PACS are the same as for MVTec. For MNIST/MNIST-M, all images are in their original scale, which are 28 × 28. We take ResNet18 \cite{he2016deep} as the backbone of the teacher encoder. 

With all datasets, the learning rate is set to 0.005 with a batch size of 16 and is optimized by Adam \cite{kingma2014adam} optimizer with $\beta=(0.5,0.999)$. The model is trained 20 epochs on MVTec, PACS, CIFAR-10 and 5 epochs on MNIST/MNIST-M dataset. The pseudo code of our training is shown in Algorithm \ref{alg:train_frame}.

\begin{algorithm}[H]
\caption{The pseudo code of our training}
\label{alg:train_frame}
\begin{algorithmic}[1]
\For{each batch ($ori, augs$) in dataloader}
\State $ens\_ori \gets encoder(ori)$ \Comment{Return a tuple with 3 embedded features from three residual encoder blocks, ordered from low-level features to abstract features}
\State $bn\_ori \gets bn(ens\_ori) $ \Comment{The feature at Bottleneck}
\State $des\_ori \gets decoder(bn\_ori) $ \Comment{Return a tuple with 3 reconstructed features from three residual decoder blocks, ordered from low-level features to abstract features}
\State $loss\_ori \gets loss(ens\_ori, des\_ori) $
\State $losses\_abs \gets 0 $
\State $losses\_lowf \gets 0 $

\For{each augmented image $aug$ in $augs$}
\State $ens\_aug \gets encoder(aug) $ 
\State $bn\_aug \gets bn(ens\_aug) $ 
\State $des\_aug \gets decoder(bn\_aug)$  
\State $loss\_abs \gets loss(bn\_ori,bn\_aug)$
\State $loss\_lowf \gets loss(des\_ori[0],des\_aug[0])$ 
\State $losses\_abs \gets losses\_abs + loss\_abs$
\State $losses\_lowf \gets losses\_lowf + loss\_lowf $
\EndFor

\State $losses\_abs \gets losses\_abs/N$
\State $losses\_lowf \gets losses\_lowf/N$
\State $sum\_loss \gets alpha\_ori \times loss\_ori + alpha\_abs \times losses\_abs + alpha\_lowf \times loss\_lowf$

\State Compute gradients of $sum\_loss$ with respect to the trainable parameters of the model
\State Update the trainable parameters of the model using the optimizer
\EndFor
\end{algorithmic}
\end{algorithm}

\subsection{Inference}
For a given test sample $x$, our test time augmentation method performs the augmentation as follows:
\begin{equation}
\textrm {FDM}(\mathcal{C} ,\mathcal{V} , \alpha ): \mathcal{C} _{\tau _i} = (1-\alpha) {\mathcal{C} _{\tau_i}} + \alpha  {\mathcal{V} _{\kappa _i}} 
  \label {eq:sumloss}, 
\end{equation}
where ${\{\mathcal{C} _{\tau _i}\}}^n_{i=1}$ and ${\{\mathcal{V} _{\kappa _i}\}}^n_{i=1}$ are sorted values of embedded feature $\mathcal{C}$ and $\mathcal{V}$ in ascending order.  Here, $n$ represents the number of elements in vector $\mathcal{C}$ and $\mathcal{V}$. Note that $\mathcal{C}$ is the embedded feature of the test sample $x$, which plays the role of carrying the appearance information. $\mathcal{V}$ is the embedded feature of a normal sample randomly sampled from the training data, carrying the style information.
In this way, the semantic information of the test sample is preserved, while its style information is pulled closer to the training data's style.

To calculate the anomaly score, we use a similar method as in RD4AD. First, we calculate the anomaly maps of the test sample $x$ at multi-level feature as follows:
\begin{equation}
    \mathcal{M}^k = 1-\mathit{sim}(\mathcal{P}^k, \mathcal{L}^k)
\end{equation}
where $\mathcal{P}^k$ and $\mathcal{L}^k$ respectively are the embedded feature and the reconstructed feature of $x$ at $k^{th}$ encoding/decoding block in our method, and $\mathit{sim}$ is a cosine similarity measure. 
Next, we increase the resolution of the feature maps $\mathcal{M}^k$ to match the input image size. To accomplish this, we use a bilinear up-sampling operation denoted as $\Psi$. We then accumulate these anomaly maps pixel-wise to generate a score map via:
\begin{equation}
     S_{AL} = \sum _{k=1}^{3}\Psi (\mathcal{M}^k) 
\end{equation}
Lastly, we choose the max value in $S_{AL}$ as the anomaly score of $x$.

\begin{figure*}
\centering
\begin{subfigure}[b]{\textwidth}
    \centering
    \includegraphics[width=\textwidth]{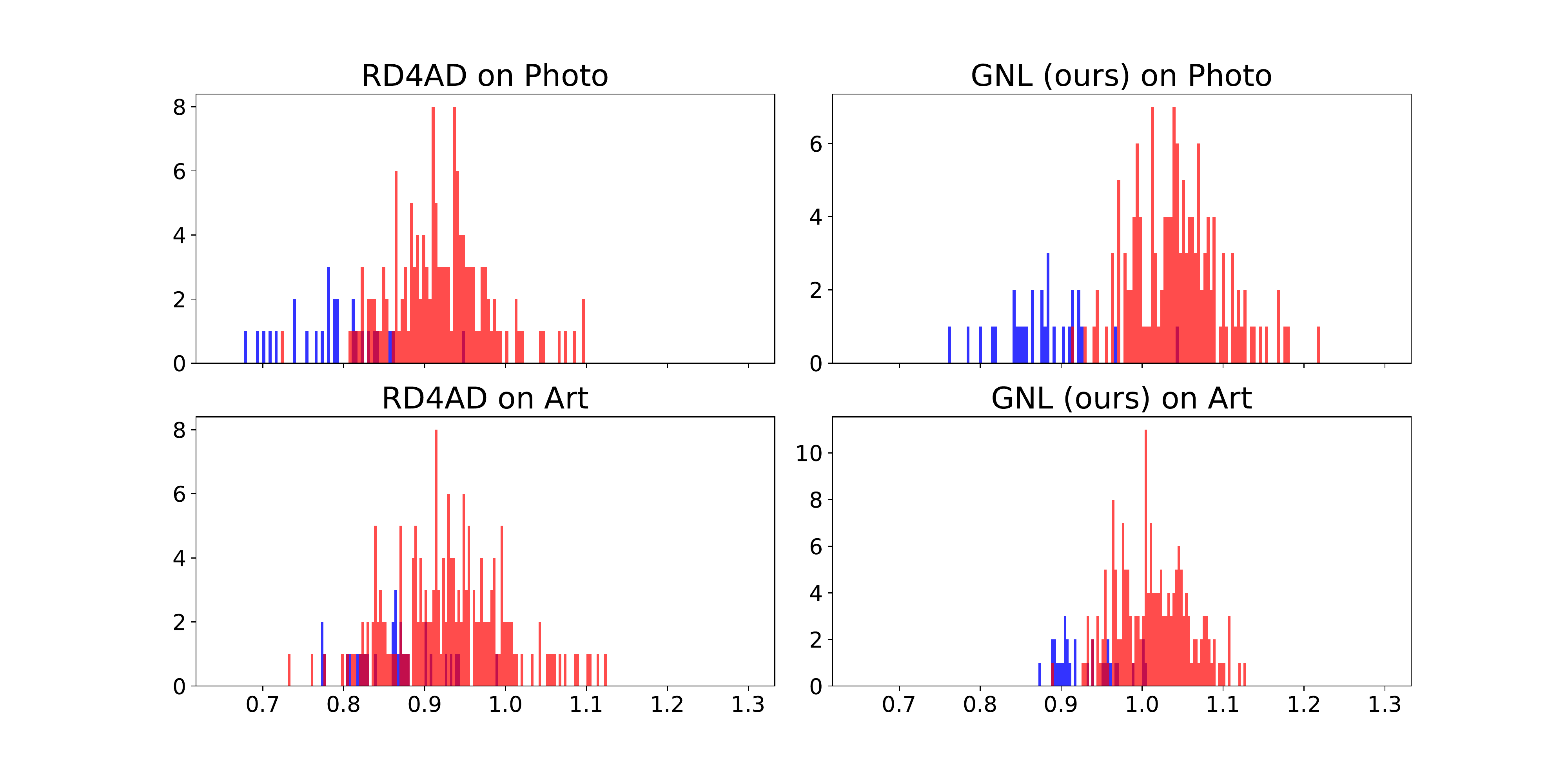}
    \caption{Anomaly scores on PACS (`elephant' as the normal class)}
    \label{fig:first}
\end{subfigure}
\vspace{0.0cm}
\begin{subfigure}[b]{\textwidth}
    \centering
\includegraphics[width=\textwidth]{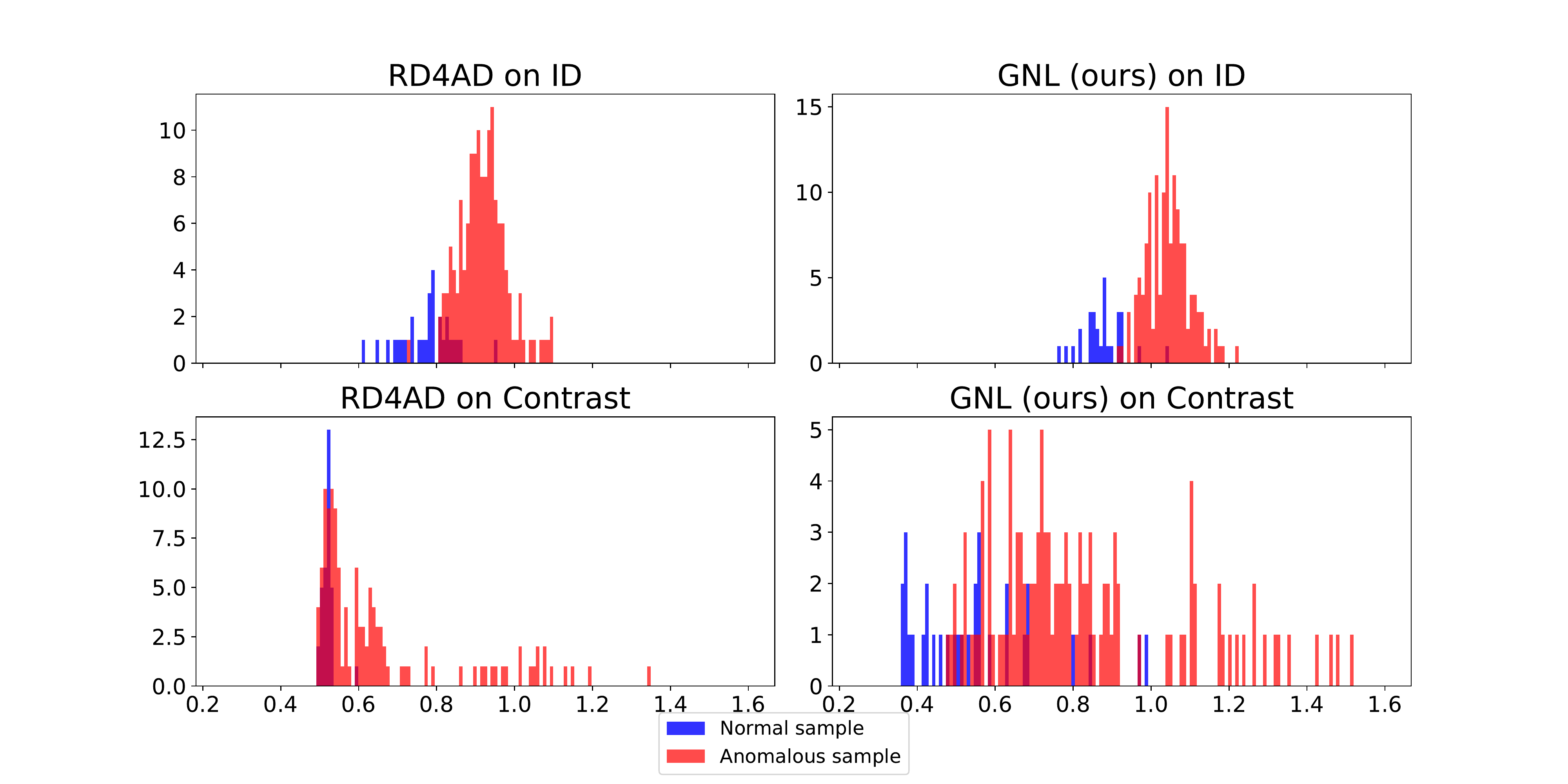}
\caption{Anomaly scores on MVTec (the `Zipper' data)}
    \label{fig:third}
\end{subfigure}
\caption{Distribution of anomaly scores yielded by our method and RD4AD. }
\label{fig:figures*}
\end{figure*}

\begin{table*}[h]
\centering
\scalebox{0.65}{
\begin{tabular}{c|c|c|c|c|c|c|c|c|c|c|c|c|c|c|c|c|c|c|c|c}
\hline
Class& \multicolumn{2}{c}{0} & \multicolumn{2}{|c}{1} & \multicolumn{2}{|c}{2} & \multicolumn{2}{|c}{3} & \multicolumn{2}{|c}{4} & \multicolumn{2}{|c}{5} & \multicolumn{2}{|c}{6} & \multicolumn{2}{|c}{7} & \multicolumn{2}{|c}{8} & \multicolumn{2}{|c}{9}  \\
\hline
&ID&OOD&ID&OOD&ID&OOD&ID&OOD&ID&OOD&ID&OOD&ID&OOD&ID&OOD&ID&OOD&ID&OOD\\
\hline
Deep-SVDD&99.24&48.08&99.72&52.96&96.53&46.28&96.61&51.92&96.48&50.36&99.27&48.10&99.76&52.97&96.56&46.28&96.61&51.92&96.48&50.36\\
f-AnoGAN&99.41&54.04&99.85&56.24&96.52&49.60&95.08&52.97&96.81&50.71&\textbf{99.35}&54.04&\textbf{99.89}&56.31&96.36&49.60&95.08&52.97&96.81&50.71\\
KDAD&99.85&58.22&\textbf{99.88}&57.15&98.42&52.99&\textbf{99.06}&55.80&\textbf{98.38}&51.95&98.33&57.11&99.49&55.51&98.69&52.02&98.45&56.88&98.16&51.12\\
RD4AD&99.56&71.50&99.50&60.09&\textbf{99.11}&51.93&98.01&55.73&96.75&50.56&98.90&58.34&99.79&64.60&\textbf{99.21}&57.37&98.90&55.77&99.17&55.03\\
\hline
Augmix&99.78&70.83&98.64&62.84&97.66&53.28&98.41&58.29&97.51&54.25&97.12&62.15&98.83&62.33&98.32&61.44&97.38&55.91&98.98&54.74\\
Mixstyle&99.61&69.90&99.57&58.54&98.76&51.19&98.85&56.04&95.83&49.10&98.92&58.43&99.57&63.14&99.16&56.74&98.99&54.98&99.17&54.14\\
EFDM&98.09&69.60&99.36&59.33&98.39&50.74&98.86&55.53&95.79&49.96&98.81&58.30&99.59&62.92&99.09&56.66&\textbf{99.02}&55.09&\textbf{99.18}&54.19\\
Jigsaw&\textbf{99.90}&71.13&99.86&60.83&99.01&51.71&98.85&56.62&96.18&52.71&98.74&59.15&99.75&64.18&99.16&58.08&98.54&55.86&99.00&54.81\\
\hline
GNL (Ours)&99.40&\textbf{80.54}&99.55&\textbf{71.95}&96.52&\textbf{63.87}&95.92&\textbf{64.69}&95.05&\textbf{64.91}&94.48&\textbf{75.33}&97.93&\textbf{79.48}&97.46&\textbf{71.25}&94.69&\textbf{64.43}&98.13&\textbf{72.27}\\

\hline
\end{tabular}
}

\caption{Full AUROC (\%) results on MNIST/MNIST-M. }
\label{tab:mvtec}
\end{table*}


\begin{table*}[h]
\centering
\scalebox{0.7}{
\begin{tabular}{c|c|c|c|c|c|c|c|c|c|c|c|c}
\hline
Dataset& \multicolumn{4}{c}{Dog} & \multicolumn{4}{|c}{Elephant} & \multicolumn{4}{|c}{Giraffe}  \\
\hline
Domain&Photo&Art&Cartoon&Sketch&Photo&Art&Cartoon&Sketch&Photo&Art&Cartoon&Sketch\\
\hline
Deep-SVDD&43.25&55.60&42.99&38.00&47.47&53.65&40.86&37.09&36.39&53.59&38.44&37.26\\

f-AnoGAN&46.30&54.06&42.90&42.90&67.36&44.09&\textbf{79.99}&34.11&64.96&51.13&47.30&51.79\\
KDAD&\textbf{76.15}&62.50&40.38&41.53&91.78&50.52&76.75&15.53&87.91&\textbf{55.67}&\textbf{65.53}&63.79\\
RD4AD&70.39&67.16&47.77&53.57&\textbf{92.07}&58.89&65.81&61.20&76.82&46.20&53.57&46.73\\
\hline
Augmix&70.36&64.33&47.08&53.80&83.36&58.83&66.31&67.16&63.75&48.38&51.44&45.70\\
Mixstyle&72.63&65.61&48.46&52.99&86.97&60.72&65.93&63.69&74.72&48.42&55.64&43.79\\
EFDM&71.81&67.06&46.95&57.05&85.46&60.80&67.12&64.61&77.64&47.21&58.27&41.96\\
Jigsaw&47.37&44.29&40.43&38.38&62.27&60.40&56.61&47.80&68.59&51.27&45.32&46.81\\
\hline
GNL (Ours)&76.13&\textbf{70.04}&\textbf{57.75}&\textbf{59.35}&90.86&\textbf{66.20}&74.83&\textbf{67.80}&\textbf{88.27}&53.80&54.21&\textbf{64.11}\\

\hline
\end{tabular}
}
\caption{Full AUROC (\%) results on PACS (Part I). }
\label{tab:mvtec}
\end{table*}

\vspace{1cm}

\begin{table*}[h]
\centering
\scalebox{0.7}{
\begin{tabular}{c|c|c|c|c|c|c|c|c|c|c|c|c|c|c|c|c}
\hline
Dataset& \multicolumn{4}{c}{Guitar} & \multicolumn{4}{|c}{Horse} & \multicolumn{4}{|c}{House} & \multicolumn{4}{|c}{Person} \\
\hline
Domain&Photo&Art&Cartoon&Sketch&Photo&Art&Cartoon&Sketch&Photo&Art&Cartoon&Sketch&Photo&Art&Cartoon&Sketch\\
\hline
Deep-SVDD&41.79&55.20&44.47&39.51&43.07&53.39&39.24&38.19&38.89&52.69&39.92&44.92&35.25&49.83&42.70&41.40\\
f-AnoGAN&42.82&34.65&56.25&\textbf{96.94}&51.72&50.40&39.76&33.28&58.76&53.17&49.47&\textbf{94.21}&97.45&63.57&51.25&93.19\\
KDAD&77.19&53.79&82.26&67.13&\textbf{85.41}&51.99&51.69&43.31&\textbf{98.76}&91.12&65.53&64.54&\textbf{100.00}&\textbf{74.39}&\textbf{56.31}&\textbf{64.00}\\
RD4AD&76.62&59.35&76.71&49.48&64.15&59.18&46.93&53.24&93.52&76.29&79.92&61.65&96.85&60.39&51.66&59.53\\
\hline
Augmix&63.68&\textbf{60.15}&67.86&55.23&64.22&56.07&48.66&46.52&95.62&74.49&80.38&79.37&93.46&61.26&51.03&57.22\\
Mixstyle&68.56&59.98&77.22&47.19&55.57&54.88&50.70&51.91&94.97&74.85&76.81&64.21&94.19&62.03&51.72&60.43\\
EFDM&65.35&56.99&77.79&47.78&61.05&56.75&52.82&53.51&92.47&73.34&80.64&63.45&95.49&61.70&51.45&61.04\\
Jigsaw&55.63&51.59&61.96&95.72&55.45&50.44&42.57&40.13&70.28&52.29&81.26&88.34&75.71&57.54&48.69&77.90\\
\hline
GNL (Ours)&\textbf{85.37}&57.68&\textbf{82.53}&45.33&77.59&\textbf{60.20}&\textbf{63.64}&\textbf{69.71}&97.55&\textbf{91.14}&\textbf{88.79}&75.42&97.95&60.31&53.97&55.01\\

\hline
\end{tabular}
}

\caption{Full AUROC (\%) results on PACS (Part II). }
\label{tab:mvtec}
\end{table*}

\begin{sidewaystable*}[t!]
\centering
\scalebox{0.57}{
\begin{tabular}{c|c|c|c|c|c|c|c|c|c|c|c|c|c|c|c|c|c|c|c|c|c|c|c|c|c}
\hline
Dataset& \multicolumn{5}{c}{Carpet} & \multicolumn{5}{|c}{Leather} & \multicolumn{5}{|c}{Grid} & \multicolumn{5}{|c}{Tile} & \multicolumn{5}{|c}{Wood}\\
\hline
Corruption&ID&Brighness&Contrast&Blur&Noise&ID&Brighness&Contrast&Blur&Noise&ID&Brighness&Contrast&Blur&Noise&ID&Brighness&Contrast&Blur&Noise&ID&Brighness&Contrast&Blur&Noise\\\hline
Deep-SVDD&54.74&33.03&53.77&44.86&49.96&64.12&35.16&38.07&63.73&39.08&84.13&91.23&74.02&85.55&76.19&74.86&41.88&45.67&70.53&68.25&89.21&75.96&54.21&89.65&76.75\\
f-AnoGAN&65.46&42.00&39.26&17.29&19.17&81.15&38.90&75.24&60.60&57.19&86.06&24.37&30.43&23.11&23.36&58.88&57.06&59.59&47.53&46.94&86.12&50.35&41.02&32.92&32.39\\
KDAD&76.59&74.70&52.68&77.44&79.44&94.29&95.41&83.22&98.99&95.55&53.38&51.54&30.74&47.65&52.77&91.70&93.81&64.86&90.45&91.87&89.24&82.49&28.51&85.23&81.17\\
RD4AD&98.75&98.30&96.35&98.60&98.18&\textbf{100.00}&\textbf{100.00}&96.31&\textbf{100.00}&\textbf{100.00}&\textbf{100.00}&\textbf{100.00}&97.63&99.92&98.19&99.09&99.46&99.22&99.37&99.76&\textbf{99.39}&\textbf{98.65}&98.66&\textbf{99.15}&\textbf{98.95}\\
\hline
Augmix&98.53&98.39&96.80&98.14&97.93&99.14&99.85&95.38&96.59&99.26&95.27&96.52&98.19&99.28&95.63&97.39&97.59&96.01&95.85&97.36&94.68&90.96&95.94&91.02&95.00\\
Mixstyle&98.68&98.30&97.00&98.58&98.45&\textbf{100.00}&\textbf{100.00}&96.55&\textbf{100.00}&\textbf{100.00}&99.58&99.28&98.36&\textbf{99.97}&\textbf{98.25}&99.25&\textbf{99.67}&99.43&99.52&99.67&99.27&98.36&97.78&98.95&98.92\\
EFDM&98.70&98.46&96.71&98.52&98.19&\textbf{100.00}&\textbf{100.00}&96.28&\textbf{100.00}&\textbf{100.00}&99.53&99.30&\textbf{98.55}&99.44&97.44&99.40&99.63&99.43&\textbf{99.65}&\textbf{99.77}&99.15&98.04&97.78&98.77&98.65\\
Jigsaw&96.91&95.02&95.21&97.33&94.60&93.77&95.70&82.86&98.30&92.32&80.20&81.70&74.94&79.50&78.89&80.52&71.76&73.51&72.01&72.13&90.38&87.31&78.33&83.95&86.81\\
\hline
GNL (Ours)&\textbf{99.48}&\textbf{99.23}&\textbf{99.20}&\textbf{99.52}&\textbf{99.11}&\textbf{100.00}&\textbf{100.00}&\textbf{99.93}&\textbf{100.00}&99.90&98.86&98.16&97.66&98.00&95.80&\textbf{99.59}&\textbf{99.67}&\textbf{99.76}&98.85&99.59&98.68&97.69&\textbf{98.77}&98.36&98.33\\
\hline
Dataset& \multicolumn{5}{c}{Bottle} & \multicolumn{5}{|c}{Hazelnut} & \multicolumn{5}{|c}{Cable} & \multicolumn{5}{|c}{Capsule} & \multicolumn{5}{|c}{Pill}\\
\hline
Deep-SVDD&78.25&60.40&59.84&89.13&76.90&84.79&31.93&16.11&85.04&51.36&73.41&50.45&47.96&73.84&70.41&58.06&62.48&68.83&64.15&47.90&64.59&39.90&36.05&60.48&49.87\\
f-AnoGAN&93.25&49.37&45.08&27.70&27.78&64.96&50.91&58.76&35.87&38.59&55.36&77.98&54.80&54.46&55.98&59.59&43.12&68.65&48.22&62.70&80.22&52.87&52.31&67.69&63.27\\
KDAD&99.71&93.76&87.43&99.55&\textbf{99.02}&98.38&94.32&88.03&97.97&95.36&90.55&81.28&53.23&91.28&91.78&80.67&81.18&51.66&80.65&74.58&78.99&73.31&49.17&79.69&75.23\\
RD4AD&\textbf{100.00}&\textbf{100.00}&98.47&\textbf{100.00}&83.33&99.96&\textbf{100.00}&99.21&99.95&99.58&96.24&96.63&90.48&95.00&96.23&97.45&88.63&81.44&95.34&76.51&\textbf{96.90}&84.08&86.05&96.12&76.44\\
\hline
Augmix&99.26&99.29&98.52&98.78&91.15&93.10&97.15&99.56&94.17&99.31&87.21&84.70&87.35&87.72&88.12&97.50&\textbf{94.86}&86.97&93.62&\textbf{87.64}&95.67&90.15&91.79&94.45&\textbf{86.58}\\
Mixstyle&\textbf{100.00}&\textbf{100.00}&\textbf{99.53}&\textbf{100.00}&81.90&99.96&99.99&99.38&99.82&99.77&96.73&96.43&90.92&95.84&96.42&97.26&89.27&83.37&95.37&76.13&96.72&86.29&86.44&96.04&78.40\\
EFDM&99.95&\textbf{100.00}&99.26&99.97&77.80&\textbf{100.00}&99.96&99.37&99.93&99.64&96.42&96.16&92.35&95.71&96.95&\textbf{97.88}&91.04&83.73&\textbf{95.84}&76.48&96.85&88.22&88.31&\textbf{96.26}&78.37\\
Jigsaw&76.54&74.47&70.82&79.36&76.15&82.29&81.08&45.57&83.67&86.65&64.59&57.96&57.01&62.53&57.96&62.60&50.99&53.05&61.34&61.84&55.21&57.95&56.15&53.79&57.42\\
\hline
GNL (Ours)&99.76&99.71&99.36&99.87&97.83&\textbf{100.00}&\textbf{100.00}&\textbf{100.00}&\textbf{100.00}&\textbf{99.87}&\textbf{96.82}&\textbf{97.29}&\textbf{97.43}&\textbf{97.23}&\textbf{97.69}&93.30&91.53&\textbf{89.65}&89.79&79.46&96.63&\textbf{90.94}&\textbf{94.41}&95.40&84.22\\
\hline
Dataset& \multicolumn{5}{c}{Transistor} & \multicolumn{5}{|c}{MetalNut} & \multicolumn{5}{|c}{Screw} & \multicolumn{5}{|c}{Toothbrush} & \multicolumn{5}{|c}{Zipper}\\
\hline
Deep-SVDD&70.79&65.42&61.58&70.96&68.29&56.06&52.10&73.12&63.83&45.45&35.23&95.08&3.18&24.37&35.66&96.94&46.94&67.22&96.11&79.44&64.47&45.80&51.42&50.11&51.16\\
f-AnoGAN&78.04&25.33&52.67&28.11&28.00&58.21&42.04&74.62&66.62&62.76&80.13&92.83&0.00&0.00&0.00&95.56&56.11&33.61&2.50&9.17&78.22&48.21&53.31&57.02&59.25\\
KDAD&88.64&86.61&64.89&90.94&87.14&81.35&86.87&79.36&80.26&84.12&73.10&89.42&77.43&56.80&35.36&92.59&85.56&55.00&93.06&91.76&93.34&86.91&94.18&92.64&\textbf{95.50}\\
RD4AD&96.34&96.17&90.58&94.36&93.97&\textbf{100.00}&\textbf{100.00}&99.43&\textbf{99.87}&92.86&98.43&97.73&95.65&97.82&84.08&99.26&89.17&97.69&99.44&98.58&97.76&98.70&84.61&97.97&55.39\\
\hline
Augmix&91.94&91.26&84.26&89.97&88.36&98.61&98.74&95.68&96.06&92.95&97.98&91.74&95.37&97.25&55.18&99.81&\textbf{96.67}&\textbf{100.00}&\textbf{100.00}&\textbf{100.00}&\textbf{98.22}&98.55&\textbf{95.79}&97.92&90.35\\
Mixstyle&95.53&95.74&88.82&93.75&93.85&\textbf{100.00}&\textbf{100.00}&99.41&\textbf{99.87}&92.83&\textbf{98.63}&\textbf{97.99}&95.44&\textbf{98.00}&85.09&98.98&88.80&\textbf{100.00}&99.91&80.56&98.18&\textbf{98.96}&84.25&\textbf{98.47}&53.59\\
EFDM&96.29&95.69&89.78&93.93&94.32&\textbf{100.00}&\textbf{100.00}&99.43&99.85&\textbf{93.53}&98.39&97.85&95.33&97.70&84.51&98.98&88.61&\textbf{100.00}&\textbf{100.00}&87.69&98.11&98.73&85.29&98.24&56.02\\
Jigsaw&69.50&68.50&64.47&66.26&69.60&60.58&65.46&65.45&65.33&51.91&53.83&68.66&64.15&60.24&61.15&81.76&85.56&68.70&80.74&73.80&60.96&58.22&67.95&63.93&67.73\\
\hline
GNL (Ours)&\textbf{97.47}&\textbf{97.00}&\textbf{95.06}&\textbf{97.58}&\textbf{96.69}&99.98&99.75&\textbf{99.90}&99.75&89.46&93.53&97.90&\textbf{96.47}&94.97&\textbf{90.11}&\textbf{100.00}&95.95&99.63&99.26&97.22&95.80&96.69&94.72&97.96&86.25\\
\hline

\end{tabular}
}

\caption{Full AUROC (\%) results on MVTec.}
\label{tab:mvtec}
\end{sidewaystable*}




\end{document}